%% file: main.tex
\documentclass[conference]{IEEEtran}
\usepackage[utf8]{inputenc}
\usepackage{url}
\usepackage{amssymb,amsmath}
\usepackage[utf8]{inputenc}
\usepackage{graphicx}
\usepackage[boxed]{algorithm2e}
\SetKw{KwBy}{by}

\newtheorem{twierdzenie}{Theorem}

\newcommand{\Bem}[1]{}
\newcommand{\Mu}{\boldsymbol\mu}

\newcommand{\totWithinSS}{SS}  
\newcommand{\wrongClustersPerc}{BC\%} 
\newcommand{\RelTotWithinSS}{relSS}  
\newcommand{\riKmeans}{$k$-means}
\newcommand{\ppKmeans}{$k$-means++}
\newcommand{\pbKmeans}{$k$-means++B}
\newcommand{\glKmeans}{glob-$k$-means}
\newcommand{\mdKmeans}{md-$k$-means}
\newcommand{\tcKmeans}{tc-$k$-means}

\title{Are Easy Data Easy (for K-Means)}

\author{
\IEEEauthorblockN{ Mieczys{\l}aw A. K{\l}opotek 
}
\IEEEauthorblockA{
 Institute of Computer Science,  Polish Academy of Sciences\\ ul. Jana Kazimierza 5, 01-248 Warsaw, Poland\\
Email: \url{klopotek@ipipan.waw.pl}
}
}

\allowdisplaybreaks
\begin{document}

\maketitle

\begin{abstract}
This paper investigates the capability of correctly recovering well-separated clusters by various brands of the $k$-means algorithm.   
The concept of well-separatedness used here is derived directly from the common definition of clusters, which imposes an interplay between the requirements of within-cluster-homogenicity and between-clusters-diversity. 
Conditions are derived for a special case of well-separated clusters such that the global minimum of $k$-means cost function coincides with the well-separatedness. 
An experimental investigation is performed to find out whether or no various brands of $k$-means are actually capable of discovering well separated clusters. It turns out that they are not.   
A new algorithm is proposed that is a variation of $k$-means++ via repeated {sub}sampling when choosing a seed. The new algorithm outperforms four other algorithms from $k$-means family on the task. 
\end{abstract}

\section{Introduction}

As visible from majority of papers (see e.g. \cite{Nikhare:2018}, \cite{Grace:2014}), 
\emph{clustering} is most frequently understood as splitting data into several subsets such that each of these clusters consists of data objects with \emph{high intra-cluster-similarity} and \emph{low inter-cluster-similarity}. 
We investigate in this paper the behaviour of algorithms from the $k$-means family under this favourable assumption. $k$-means is one of the most famous clustering algorithms invented  more than half a century. 
Thanks to a multitude  optimization and enhancement works, $k$-means  is also  the most popular clustering algorithm in use  in various domains. 
For recent overviews of various brands of this algorithm see e.g. 
\cite{Ashabi:2021} or  \cite{STWMAKSpringer:2018}.

If we take   seriously the assumption of {high intra-cluster-similarity} and {low inter-cluster-similarity}, then the distance of a {data}point  to other {data}points of the same cluster should be lower than the distance of the same {data}point to {data}points of other clusters. This assumption fits the definition of so-called \emph{nice clustering} \cite{AckermanD:2014}.  
If we push further and require that the minimum
separation between clusters be larger than the maximum cluster diameter, then we speak about \emph{perfect clustering} \cite{AckermanD:2014}. 
If we extend this requirement not only to {data}points but to the entire {hyper}balls centered at cluster gravity center and encompassing all cluster {data}points, then we speak about \emph{perfect-ball-clustering}   \cite{RAK:MAK:2019:perfectball}.
Provably, incremental $k$-means algorithm introduced in \cite{AckermanD:2014}, can discover the perfect ball clustering if it exists \cite{RAK:MAK:2019:perfectball} and if we know the correct number of clusters in advance. It shall be stressed, however, that incremental $k$-means does not seek to optimize the traditional $k$-means cluster quality function.   


This paper presents a study of $k$-means properties when dealing with this most simplistic case of clusters. The  merit of studying such a trivial case of clustering is two-fold. First, the concept of a cluster is not so clear in general, as one may expect. So the study of clustering should start with conceptual investigation of the case when the concept seems to be clear-cut. Second, it turns out that the well-studied and widely applied $k$-means algorithm may fail under these ideal circumstances. We point at an algorithm based on $k$-means that is well suited under such circumstances. We hope that it may constitute a hint for elaboration of some in-between versions for leaning $k$-means towards human cluster expectations.  

As the $k$-means family of clustering algorithms is astoundingly  large, 
\cite{Ashabi:2021},  \cite{STWMAKSpringer:2018}, we have selected only a few representatives for this study: the traditional $k$-means \cite{Lloyd:1982},  
 $k$-means++   \cite{CLU:AV07}
\cite{Song:2010}, "most distant initialization" algorithm  \cite{corr/abs-1903-10025},  global $k$ means (GKM) \cite{Likas:2003:GKM}  and our own variation. 
These algorithms seem to be representative for diverse ideas on algorithm initialization which is essential for well-separated data (as this study confirms).

\section{Previous work}

The widely used 
$k$-means clustering algorithm  (\cite{Lloyd:1982}, for various versions see e.g. \cite{STWMAKSpringer:2018})
seeks, for a dataset $\mathbf{X}$, to minimize the function
\begin{equation} \label{eq:Q::kmeans}
Q(\Gamma)=\sum_{i=1}^m\sum_{j=1}^k u_{ij}\|\textbf{x}_i - \boldsymbol{\mu}_j\|^2
=\frac12\sum_{j=1}^k \frac{1}{n_j} \sum_{\mathbf{x}_i    \in C_j}\sum_{  \mathbf{x}_l \in C_j} \|\mathbf{x}_i - \mathbf{x}_l\|^2 
\end{equation} 
\noindent 
under some partition $\Gamma$ into 
 $k$  clusters, 
where  $u_{ij}$ is an indicator (%
$u_{ij}\in\{0,1\}$) of the membership of data point $\textbf{x}_i$ in the cluster $C_j$ having the center at $\boldsymbol{\mu}_j$. 

We will call   \emph{$k$-means-ideal} such an algorithm that finds a $\Gamma_{opt}$ that attains the minimum of function $Q(\Gamma)$. 
It is known that it is a hard task.

Many researchers try to  approximate 
$k$-means-ideal within a reasonable error bound (e.g. $9+\epsilon$ by \cite{Kanungo:2002}),  via cleverly initiated $k$-means type algorithms, (e.g. $k$-means++  
 \cite{CLU:AV07}
\cite{Song:2010}, or "most distant initialization" algorithm which is claimed to be nearly as effective as   $k$-means++ \cite{corr/abs-1903-10025}),  
or other algorithms, like (robust) single link, 
or   approximating the quality function value by using a higher $k'$ when clustering. 
Recently, attention of the researchers was attracted by the so-called global $k$ means (GKM) \cite{Likas:2003:GKM} due to its successes in finding quite well initial seeds for $k$-means in a more deterministic way than in the other algorithms, which has a number of follow-ups aiming at improving its efficiency  (see e.g. \cite{Xie:2011:GKM}, \cite{Bagirov:2011:GKM} , \cite{Agrawal:2013:GKM} and many other). 

The $k$-means like algorithms are of interest not only because their efficiency, but also due to interesting theoretical property like consistency in the limit for growing sample sizes \cite{MAK:2020:consistencykpp},
applicability of kernel-trick for Euclidean \cite{MAK:2019:trickEuclid} and non-Euclidean space \cite{MAK:2020:trickNonE}, 
existence of cluster-preserving transformations 
\cite{MAK:2022:clusterpres}, possibility to check clusterability \cite{MAK:2017:aposteriori},  
applicability in label-free test set evaluations \cite{miao2023kmeans}, privacy preserving \cite{WANG2023120396}
and many other \cite{MAK:2023MachLearn}, \cite{MAK:2020:shape}, \cite{MAK:2019:richness}. 
Therefore, intense studies of $k$-means family are vital.

The $k$-means clustering of well-separated data is a topic studied in the context of diverse applications \cite{Braverman:2011}, \cite{Ostrovsky:2013}. 
In various publications (including \cite{Ostrovsky:2013}), 
assumptions are possibly made about the structure of the data, e.g. about sufficiently large gaps between the clusters, or low variance compared to cluster center distances etc. 
We will pursue here this line of research by investigating the capability of $k$-means like algorithm to discover well separated clusters. 

In this paper, we will use $k$-means implementation in $R$, which 
allows to investigate several brands of algorithms from this family differing on the way it is initialized.

\section{The Easy Case for $k$-Means}

Recall that $k$-means is intended to cluster a dataset $D$ into the set $\Gamma$ of $k$ non-empty non-intersecting groups, called clusters such that they should reach the minimum of the $k$-means cost function, Eq. (\ref{eq:Q::kmeans}), that may be reformulated as:
$$Q(\Gamma,d)=\sum_{C\in \Gamma}  \sum_{i\in C} \sum_{j\in C} d^2(i,\Mu(C)) $$
where $\Mu(C)$ is the position of the center of gravity of cluster C, and $d$ is the (Euclidean) distance induced by the embedding of $D$ in some $\mathbb{R}^m$. 
The ideal-$k$-means is one that reaches this minimum, by discovering $\Gamma_{true}$ (but with prohibitive execution time), while there exist multiple implementations, real-$k$-means algorithms that find some $\Gamma$ approximating this goal only (sticking in some local minimum of $Q(\Gamma$)

Consider a dataset $D$,
and its intended clustering $\Gamma_I$ into $k$ clusters such that each $C\in \Gamma$ has the same cardinality ($|C|=n$). We seek for conditions under which $\Gamma_I=\Gamma_{true}$. To this end, we assume that each cluster $C\in \Gamma_I$ can be enclosed into a (hyper)ball of radius $R$ with center at $\Mu(C)$, for some fixed $R$. Furthermore there exist a $g$ such that distances between various mentioned ball surfaces amount to at least $g$. We will impose restrictions on $g$. 

Under these circumstances consider a clustering $\Gamma$ of $D$, obtained by some real $k$-means. The first possibility is that each $\Mu(C)$ for each $C$ in $\Gamma$ lies outside of each $C_I\in \Gamma_I$ at a distance over $R$ from the surface. 
In this case $Q(\Gamma_I,d)<Q(\Gamma,d)$, hence $\Gamma$ is not optimal.

Let us discuss other case that is  
there exists at least one cluster $C_N\in \Gamma_I$ such that some $\Mu(C_1)$, $C_1\in\Gamma$ is at the distance at most $R$ from the ball of $C_N$. This means that for each $e\in C_N$: $d(e,\Mu(C_1))\le 3R$. 
Let $z \in C_2 \cap C_N$ where $C_T\in \Gamma_I, C_T \ne C_N$ and there exists a $C_2\in \Gamma$ such that $\Mu(C_2)$ is at most $R$ away from $C_T$ ball.  
Then $d(z,\Mu(C_1))>g-R$. If $g>4R$, then  $d(z,\Mu(C_1))>3R\ge d(z,\Mu(C_2))$ and therefore 
for $\Gamma'$ such that 
$\Gamma'=(\Gamma/\{C_1,C_2\})
\cup\{C_1/\{z\},C_2\cup\{z\}\}$
we have 
$Q(\Gamma',d)<Q(\Gamma,d)$, hence $\Gamma$ is not optimal.

So let us discuss the case excluding the above that implies the situation  where  for each $C \in \Gamma$ such that if $\Mu(C)$ is at most $R$ away from some ball of $C_N\in \Gamma_I$, then each element $e$ of $C_N$ for each $C_2\in \Gamma$: $d(e,\Mu(C_2))>3R$.  

We can have two cases: either each $C_N\in\Gamma_I$ has exactly one $C\in \Gamma$ such that $\Mu(C)$ is at most $R$ away from $C_N$ ball
or
some  $C_N\in\Gamma_I$ have at least  one $C\in \Gamma$ such that $\Mu(C)$ is at most $R$ away from $C_N$ ball and other do not. 

In the first case obviously $Q(\Gamma_I,d)\le Q(\Gamma,d)$ because each element of a close cluster that is not in the intended cluster to which it is close, is further away from the cluster center than if it would be if moved to the close cluster to its intended cluster.

In the second  let us distinguish close $\Mu(C)$'s as such that are at most $R$ away from some intended cluster ball
(and the respective $C$s will be called close clusters) 
and the other $\Mu(C)$'s that shall be called "remote"
(and the respective $C$s will be called remote clusters).
An intended cluster ball with a $\Mu(C)$ not further than $R$ away from it will be called "with satellite", while the other "without satellite".  

Let us denote the set of all elements that belong to a close cluster but do not belong to the respective intended cluster with satellite with $A_c$. 
For each element $e$ of $A_c$, belonging to some $C_T\in \Gamma_I$ and some $C\in \Gamma$ its distance to $\Mu(C)$ is at least $g-R$, and its distances to all the other cluster centers of $\Gamma$ amount to at least to $g-3R$. All other elements of $C_T$ have distances to cluster centers of $\Gamma$ not smaller than $g-3R$. There exists at least one such cluster $C_T$. 
There exist clearly at most $k-1$ close clusters. If we switch to $\Gamma_I$ from $\Gamma$, then in these clusters the overall contribution to $Q$ may increase at most by $R^2(k-1)n$.
The contribution of the mentioned remote cluster $C_T$ (there exists at least one such cluster) will decrease at least by $(g-3R)^2n$. For all the other remote clusters the contribution will for sure not increase. 
Therefore, if we are interested in having $Q(\Gamma_I,d)<Q(\Gamma,d)$, it will be sufficient to have:  
$R^2(k-1)n< (g-3R)^2 n$. 
Hence 
$R^2(k-1)< (g-3R)^2 $. 
$R\sqrt{k-1}<g-3R$
$R(\sqrt{k-1}+3)<g$.
So two conditions have to hold: 
$g>4R$ and $g>R(\sqrt{k-1}+3)$. The first is superfluous as usually we have at least two clusters to consider. 

In 2D on the grid, there are not $k-1$ but rather only at most 8 neighbours of $C_T$ to consider. So we will be working with $g>R(\sqrt{8}+3). $

Out of these considerations, we can formulate theorems:

\begin{twierdzenie}\label{thm:wellsepclusters}
    For a clustering $\Gamma$ (embedded in Euclidean space) such that each cluster can be enclosed in a ball with radius $R$ centered at its gravity center and the distances between the enclosing balls are at least $R(\sqrt{k-1}+3)$, where $k=|\Gamma|$m, then $\Gamma$ minimizes the cost function (\ref{eq:Q::kmeans}). 
\end{twierdzenie}
\begin{twierdzenie} \label{thm:wellsepclusters2D}
If the clustering from the previous theorem is embedding in two-dimensional space, then it is sufficient that the distances between enclosing balls are not smaller than $R(\sqrt{8}+3)$
\end{twierdzenie}
    
\section{Experiments} 
The goal of our experiments is to see whether or not various brands of $k$-means can discover the optimal clustering for well separated data (easy conditions). 
We investigate clusters lying on a regular grid, clusters slightly dislocated from regular grid, data without noise and data with noise. Various sizes of clusters will be investigated.

\subsection{Data}

Two-dimensional synthetic data will be used. The minimum distance between clusters is taken from Theorem \ref{thm:wellsepclusters2D}. 
All clusters will be of the same size. In various experiments this size will vary from 12 to 06. The number of clusters vary from 25 to 100. 

Each element of a cluster is generated as follows: From the intended cluster center take a random direction (uniform sampling) and then sample uniformly the distance from intended cluster center (from zero to cluster radius). 
The noise is generated in the same way, with the exception that the distance from the intended cluster center is radius plus sample from Gaussian distribution (with standard deviation equal cluster radius). 

All cluster radii are identical. Clusters are placed on regular grids. Distortion of cluster center position is taken uniformly from, the assumed range. 

Synthetic data is used because to the best of our knowledge no real data sets with guaranteed global minimum of $k$-means cost function exist (only data{}sets are known for which no better solution was found in spite of computational effort).

\subsection{Methods}

\begin{algorithm}
\KwData{$D$ - a set of objects embedded in Euclidean space (that is for each $i\in D$ there exists its representation $\mathbf{x_i}$ in Euclidean space), to be clustered\\
$k$ - the number of clusters to be returned  }
\KwResult{ 
$\Gamma$ - the clustering of $D$ into $k$ clusters }
 Initialize the set $M$ of $k$ cluster centers - step depends of the concrete algorithm\; 
\While{termination not reached} {
 Create a clustering $\Gamma$ assigning to each $\Mu\in M$ the subset of $D$ of elements closest to $\Mu$ ;
Drop $M$ and create a new one such that for each cluster $C\in \Gamma$ compute $\Mu(C)= \frac{1}{|C|} \sum_{i\in C} \mathbf{x_i} $\; 
} 
\Return $\Gamma$\;
\caption{
The framework of the various brands of $k$-means algorithms; termination condition may be a fixed number of loop runs or no change of clustering within a loop run.
}\label{alg:kmeansFrame}
\end{algorithm}

\begin{algorithm}
\KwData{$D$ - a set of objects embedded in Euclidean space, to be clustered\\
$k$ - the number of clusters to be created  }
\KwResult{ 
$M$ - the set of initial centroids to be used in the $\riKmeans$}
Pick $M$ as a random sample of $k$ elements from $D$ without repetition\;
\Return $M$\;
\caption{
The initialization for the \riKmeans{} algorithm, used in conjunction with Algorithm \ref{alg:kmeansFrame}.
}\label{alg:rikmeansInit}
\end{algorithm}

\begin{algorithm}
\KwData{$\Gamma$ - the intended clustering\\
$k$ - the number of clusters to be created  ($|\Gamma|=k$ required }
\KwResult{ 
$M$ - the set of initial centroids to be used in the $\tcKmeans$}
Create $M$  such that for each cluster $C\in \Gamma$ compute $\Mu(C)= \frac{1}{|C|} \sum_{i\in C} \mathbf{x_i} $\; 
\Return $M$;
\caption{
The initialization for the \tcKmeans{} algorithm, used in conjunction with Algorithm \ref{alg:kmeansFrame}.
}\label{alg:tckmeansInit}
\end{algorithm}

\begin{algorithm}
\KwData{$D$ - a set of objects embedded in Euclidean space, to be clustered\\
$k$ - the number of clusters to be created  }
\KwResult{ 
$M$ - the set of initial centroids to be used in the $\riKmeans$}
Pick one element of $D$ at random and initialize $M$ with it. \;
\For{$j\leftarrow 2$ \KwTo $k$ \KwBy $1$} 
{ Assign to each element of $e\in D$ a probability $p_e$, proportional to the squared distance of $e$ to the closest element of $M$\;
Sample one element $c\in D$ according to the above-mentioned probability; 
$M \leftarrow=M \cup \{c\}$
}
\Return $M$\;
\caption{
The initialization for the \ppKmeans{} algorithm, used in conjunction with Algorithm \ref{alg:kmeansFrame}.
}\label{alg:ppkmeansInit}
\end{algorithm}

\begin{algorithm}
\KwData{$D$ - a set of objects embedded in Euclidean space, to be clustered\\
$k$ - the number of clusters to be created  }
\KwResult{ 
$M$ - the set of initial centroids to be used in the $\riKmeans$}
Pick one element of $D$ at random and initialize $M$ with it. \;
\For{$j\leftarrow 2$ \KwTo $k$ \KwBy $1$} 
{ Find the element  $c\in D$ with the highest distance to the closest element of $M$. \;
$M \leftarrow=M \cup \{c\}$
}
\Return $M$\;
\caption{
The initialization for the \mdKmeans{} algorithm, used in conjunction with Algorithm \ref{alg:kmeansFrame}.
}\label{alg:mdkmeansInit}
\end{algorithm}

\begin{algorithm}
\KwData{$D$ - a set of objects embedded in Euclidean space, to be clustered\\
$k$ - the number of clusters to be created \\
$b$ - boosting parameter}
\KwResult{ 
$M$ - the set of initial centroids to be used in the $\riKmeans$}
Pick one element of $D$ at random and initialize $M$ with it. 
\For{$j\leftarrow 2$ \KwTo $k$ \KwBy $1$} 
{ Assign to each element of $e\in D$ a probability $p_e$, proportional to the squared distance of $e$ to the closest element of $M$\;
Sample $b$ elements $B\subset D$ according to the above-mentioned probability; 
Select the element $c\in B$ for which the sum of distances of all elements of $D$ to the closest element from the set $M \cup \{c\}$ is the lowest\; 
$M \leftarrow=M \cup \{c\}$\;
}
\Return $M$\;
\caption{
The initialization for the \pbKmeans{} algorithm, used in conjunction with Algorithm \ref{alg:kmeansFrame}.
}\label{alg:pbkmeansInit}
\end{algorithm}

\begin{algorithm}
\KwData{$D$ - a set of objects embedded in Euclidean space, to be clustered\\
$k$ - the number of clusters to be created  }
\KwResult{ 
$M$ - the set of initial centroids to be used in the $\riKmeans$}
Initialize $M$ with the centroid of $D$\; 
\For{$j\leftarrow 2$ \KwTo $k$ \KwBy $1$} 
{
For each $e\in D$ compute the statistics 
$v_e = \sum_{f \in D} \max\left(0,  d^2(f,M) - d^2(f,e)\right)$\$;
Select the element $c\in D$ such that $v_c$ takes on the maximum value (maximum reduction of errors)\; 
$M \leftarrow M \cup \{c\}$\;
} 
\Return $M$\;
\caption{
The initialization for the \glKmeans{} algorithm, used in conjunction with Algorithm \ref{alg:kmeansFrame}.
}\label{alg:glkmeansInit}
\end{algorithm}

We will subsequently investigate the following versions of $k$-means:
\begin{itemize}
    \item the original $k$-means with random initialization (\riKmeans),\cite{Lloyd:1982}, see Algorithm \ref{alg:rikmeansInit}. 
    \item the $k$-means++ with probabilistic seeding with probabilities proportional to squared distances to already selected seeds (\ppKmeans), (\cite{CLU:AV07}), see Algorithm \ref{alg:ppkmeansInit}.  
    \item a boosted version of the $k$-means++, proposed in this paper (\pbKmeans), see Algorithm \ref{alg:pbkmeansInit}; the parameter $b$ is set to 15 through all experiments. 
    \item global $k$-means (\glKmeans, \cite{Xie:2011:GKM}, \cite{Bagirov:2011:GKM}, \cite{Agrawal:2013:GKM}), see Algorithm \ref{alg:glkmeansInit}. 
    \item most distant seed   $k$-means (\mdKmeans), initialized by choosing the seed most distant to existing seeds, \cite{corr/abs-1903-10025},  see Algorithm \ref{alg:mdkmeansInit}
    \item unrealistic "true center" $k$-means, seeded with true cluster centers (\tcKmeans),  see Algorithm \ref{alg:tckmeansInit}. 
\end{itemize}

\input TABLES/run-30-5-5-0-10-40-0

\input TABLES/run-30-6-6-0-10-40-0

\input TABLES/run-30-7-7-0-10-40-0

\input TABLES/run-30-8-8-0-10-40-0

\input TABLES/run-30-9-9-0-10-40-0

\input TABLES/run-30-10-10-0-10-40-0

\subsection{Evaluation}
The quality of the algorithms is evaluated as follows.
Thirty times a data sample is generated, eventually with addition of noise. 
Then each algorithm is run of the same data sample.  
The following quantities are computed:
\totWithinSS{} - total sum of squared distances between elements and the discovered cluster centers, \wrongClustersPerc{} - the percentage of intended clusters, that were not discovered by the algorithm (only regular data points are considered, noise is neglected), 
  \RelTotWithinSS{} - the relative \totWithinSS, that is \totWithinSS{} of a given algorithm divided by the lowest \totWithinSS{} among algorithms for a given dataset. 

  The presented tables contain + for each of the above three criteria + the average over 30 data{}sets (30 runs) and the standard deviation (column denoted as SD\footnote{NaN as standard deviation means that it was too close to zero, so that rounding errors led to imaginary values}). 
Additionally, the worst cases (with the highest \wrongClustersPerc)  are stored. Selected  worst cases are presented in the accompanying figures. The discovered clusters differ in color and mark shapes. The erroneous discovered clusters are mark with big black circle.

\subsection{Clusters on regular grid without noise}

\input PDFS/worst_riKmeans_10-10-0-10-40-0

\input PDFS/worst_ppKmeans_10-10-0-10-40-0

The results of the experiment in discovering correctly clusters lying on a regular grid without noise are presented in tables 
\ref{tab:5-5-0-10-40-0},
\ref{tab:6-6-0-10-40-0},
\ref{tab:7-7-0-10-40-0},
\ref{tab:8-8-0-10-40-0},
\ref{tab:9-9-0-10-40-0} and 
\ref{tab:10-10-0-10-40-0} for $k=25,36,49,64,81,100$ resp. As visible, all the algorithms detect the true clustering except for \riKmeans{} and \ppKmeans. 
The \ppKmeans{} performs significantly better than the baseline \riKmeans, and the percentage of errors of \ppKmeans{}  increases insignificantly when increasing the number of clusters by the factor of four, while \riKmeans{}  increases the error rate by 50\%. 
Sample worst cases for the  algorithms can be seen in figures \ref{fig:riKmeans-10-10-0-10-40-0} and \ref{fig:ppKmeans-10-10-0-10-40-0}. The circles indicate the erroneous clusters. Each circle is centered at the gravity center of the detected cluster. So if two circles overlap, then it means that a true cluster was split into two (or more) clusters  (there are six such cases in  figures \ref{fig:riKmeans-10-10-0-10-40-0}).. If a circle touches two true clusters then it means that the algorithm failed to recognize that there are two clusters (six such cases in  figures \ref{fig:riKmeans-10-10-0-10-40-0}).  

.

\subsection{Clusters on regular grid with noise}

\input TABLES/run-30-8-8-10-10-40-0

\input TABLES/run-30-8-8-20-10-40-0

\input TABLES/run-30-8-8-30-10-40-0

\input TABLES/run-30-8-8-40-10-40-0

\input TABLES/run-30-8-8-50-10-40-0

\input PDFS/worst_riKmeans_8-8-30-10-40-0

\input PDFS/worst_mdKmeans_8-8-30-10-40-0

\input PDFS/worst_ppKmeans_8-8-30-10-40-0

\input PDFS/worst_glKmeans_8-8-30-10-40-0

The results of clustering on regular grid with noise ranging from 10 to 50\% can be seen in tables 
\ref{tab:8-8-10-10-40-0} - 
\ref{tab:8-8-50-10-40-0}. 
All algorithms commit errors except for \tcKmeans{} and \pbKmeans. 
Worst cases are presented in figures
\ref{fig:riKmeans-8-8-30-10-40-0} - \ref{fig:glKmeans-8-8-30-10-40-0}.
The algorithms seem not to be much sensitive to noise (that is their error rate does not increase with noise much), in particular the \riKmeans{} and \ppKmeans. 
\glKmeans{}, which appears to perform worse than \mdKmeans{} in the experiments, nevertheless exhibits error stability, while \mdKmeans{} error increases systematically.

\subsection{Clusters with positions disturbed with respect to a regular grid}

The results of clustering on regular grid with noise and with dislocating clusters from grid points by 0.5 to 4 cluster radii  can be seen in tables 
\ref{tab:8-8-30-10-40-5} - 
\ref{tab:8-8-30-10-40-40}. 
Algorithms \riKmeans, \ppKmeans{} and \glKmeans{} commit errors. 
Worst cases are presented in figures
\ref{fig:riKmeans-8-8-30-10-40-40} - \ref{fig:glKmeans-8-8-30-10-40-40}.

\input TABLES/run-30-8-8-30-10-40-5

\input TABLES/run-30-8-8-30-10-40-10

\input TABLES/run-30-8-8-30-10-40-20

\input TABLES/run-30-8-8-30-10-40-40

One gets the impression that dislocation of clusters from grid positions improves the performance of all algorithms except for \riKmeans. 
This improvement stems from the fact that allowing for dislocation enforces larger gaps between clusters so that the minimum gap from Theorem \ref{thm:wellsepclusters2D} is kept. This distance increase has however bad effects on \riKmeans{} because it reduces its capability of recovery after a faulty seed initialization which is always the poorest among the algorithms.

\input PDFS/worst_riKmeans_8-8-30-10-40-40

\input PDFS/worst_ppKmeans_8-8-30-10-40-40

\input PDFS/worst_glKmeans_8-8-30-10-40-40

\subsection{Impact of cluster size on clustering}

The impact  of cluster size on clustering can be seen in tables 
\ref{tab:8-8-30-10-12-10} - 
\ref{tab:8-8-30-10-96-10}. 
Here, the cluster sizes range from 12 to 96. 
All algorithms commit errors except for \tcKmeans{} and \pbKmeans. 
Worst cases are presented in figures
\ref{fig:riKmeans-8-8-30-10-96-10} - \ref{fig:glKmeans-8-8-30-10-96-10}. 
No obvious pattern can be seen for error dependence on
cluster size. Well-separatedness plays a dominant role.

\input TABLES/run-30-8-8-30-10-12-10

\input TABLES/run-30-8-8-30-10-24-10

\input TABLES/run-30-8-8-30-10-48-10

Summarizing, we see that apparently the \pbKmeans{} algorithm demonstrated best performance among the investigated algorithms (\tcKmeans{} cannot be considered as a competitive algorithm because of the unrealistic assumption of knowing true cluster centers in advance). 

As expected, the \riKmeans{} algorithm proved to have the poorest performance. A big surprise is the quite well performance of \mdKmeans{} in spite of its simplicity. \glKmeans{} was a bit dismaying. Though it outperformed always \ppKmeans, its execution time (not reported here) was really high, several times higher than of all the other algorithms together. 
It may be seen as a surprise that \ppKmeans{} was beaten  by the simplistic \mdKmeans, and at the same time its boosted version \pbKmeans{} performed best. 
Generally, the basic weakness of $k$-means type algorithm is the need to hit each intrinsic cluster if they are well separated, because otherwise there is nearly no chance for a recovery in a later process. \ppKmeans{} represents the nice idea that large enough clusters lying far away from clusters hit so far have a high probability of hitting even if their members may have lower hitting probability than some outliers (noise) and the members of clusters already hit. But when we come close to selecting the last cluster seed, this advantage drops so that \ppKmeans{} does not perform well with a large number of clusters. This problem is at least partially overcome in our modification, that is in the \pbKmeans{} algorithm. Checking multiple seed candidates at  each step decreases the chance that an outlier or a member of hit cluster will be selected as the next seed and hence the good performance of our method. The only problem is that $b$ needs to be adjusted experimentally, while a clear relation between it and the noise level etc. would be very helpful for setting it in advance. 
\input TABLES/run-30-8-8-30-10-96-10

\input PDFS/worst_riKmeans_8-8-30-10-96-10
Some insights can be gained from the study of worst cases. One could have expected that the most problematic area would be the grid border. However, it does not seem to be so. Besides, \glKmeans{} seems to commit errors close to the grid center. So possibly, improvements of this algorithm should concentrate around the clusters close to the gravity center of all the data. 
\ppKmeans seems to concentrate clusters without a hit in other area than those hit twice, even in the absence of noise. %
A review of the \RelTotWithinSS{} in the tables shows that \riKmeans{} can increase the \totWithinSS{} even by a factor of 10. So its application in case of relatively clear cluster structure shall not be recommended. 

\input PDFS/worst_ppKmeans_8-8-30-10-96-10

\input PDFS/worst_glKmeans_8-8-30-10-96-10

\section{Conclusions}
In this research we were interesting in the question how well the popular family of algorithms, $k$-means, can handle well separated data under favourable conditions of coinciding global minimum of $k$-means cost function with the well-separation. 
We studied four popular algorithms (\riKmeans, \mdKmeans, \ppKmeans and \glKmeans) and our own algorithm \pbKmeans. 
For this purpose, we have identified the conditions under which well-separation coincides for sure with the $k$-means cost function minimum (see Theorems \ref{thm:wellsepclusters}, \ref{thm:wellsepclusters2D}.
Given this theoretical result, series of experiments of algorithm performance with and without noise, with displacement and for varying cluster sizes were performed. 
It turned out that our algorithm exhibits the best performance while having acceptable executions time. 
A further research on properties of the new algorithm may be needed, in particular on the relationship between the boosting parameter and the noise levels. Also research on real datasets with artificially added noise may be insightful. 

\bibliographystyle{plain}

\input main.bbl
\end{document}

%% file: TABLES/run-30-5-5-0-10-40-0.tex
\begin{table}
\caption{Misclustering errors for the number of clusters $k$= 5 x 5 = 25  
noise level= 0 \%, cluster radius= 10 , cluster size= 40 , 
motion freedom of clusters= 0 , minimal distance between clusters: 5.83 *R
} \label{tab:5-5-0-10-40-0} 
\begin{tabular}{|l|r|r|r|r|r|r|}
\hline
& \totWithinSS &sd& \wrongClustersPerc &sd& \RelTotWithinSS &sd\\ 
 \hline
\riKmeans& 245190.2 & 73349.84& 21.07 & 7.37& 7.57 & 2.3\\ 
 \hline
\tcKmeans& 32429.34 & 899.34& 0 & 0& 1 & 0\\ 
 \hline
\mdKmeans& 32429.34 & 899.34& 0 & 0& 1 & 0\\ 
 \hline
\glKmeans& 32429.34 & 899.34& 0 & 0& 1 & 0\\ 
 \hline
\ppKmeans& 93007.96 & 74481.14& 6 & 7.43& 2.87 & 2.31\\ 
 \hline
\pbKmeans& 32429.34 & 899.34& 0 & 0& 1 & 0\\ 
 \hline

\end{tabular}
 \end{table}

%% file: TABLES/run-30-6-6-0-10-40-0.tex
\begin{table}
\caption{Misclustering errors for the number of clusters $k$= 6 x 6 = 36  
noise level= 0 \%, cluster radius= 10 , cluster size= 40 , 
motion freedom of clusters= 0 , minimal distance between clusters: 5.83 *R
} \label{tab:6-6-0-10-40-0} 
\begin{tabular}{|l|r|r|r|r|r|r|}
\hline
& \totWithinSS &sd& \wrongClustersPerc &sd& \RelTotWithinSS &sd\\ 
 \hline
\riKmeans& 418367.7 & 67462.19& 24.35 & 4.46& 8.93 & 1.48\\ 
 \hline
\tcKmeans& 46884.35 & 1036.28& 0 & 0& 1 & 0\\ 
 \hline
\mdKmeans& 46884.35 & 1036.28& 0 & 0& 1 & 0\\ 
 \hline
\glKmeans& 46884.35 & 1036.28& 0 & 0& 1 & 0\\ 
 \hline
\ppKmeans& 140119.3 & 101557.5& 6.39 & 7.03& 3 & 2.18\\ 
 \hline
\pbKmeans& 46884.35 & 1036.28& 0 & 0& 1 & 0\\ 
 \hline

\end{tabular}
 \end{table}

%% file: TABLES/run-30-7-7-0-10-40-0.tex
\begin{table}
\caption{Misclustering errors for the number of clusters $k$= 7 x 7 = 49  
noise level= 0 \%, cluster radius= 10 , cluster size= 40 , 
motion freedom of clusters= 0 , minimal distance between clusters: 5.83 *R
} \label{tab:7-7-0-10-40-0} 
\begin{tabular}{|l|r|r|r|r|r|r|}
\hline
& \totWithinSS &sd& \wrongClustersPerc &sd& \RelTotWithinSS &sd\\ 
 \hline
\riKmeans& 608395.6 & 84216.04& 26.33 & 3.74& 9.51 & 1.38\\ 
 \hline
\tcKmeans& 64029.5 & 1054.86& 0 & 0& 1 & 0\\ 
 \hline
\mdKmeans& 64029.5 & 1054.86& 0 & 0& 1 & 0\\ 
 \hline
\glKmeans& 64029.5 & 1054.86& 0 & 0& 1 & 0\\ 
 \hline
\ppKmeans& 209584.8 & 96739.58& 7.34 & 4.85& 3.27 & 1.5\\ 
 \hline
\pbKmeans& 64029.5 & 1054.86& 0 & 0& 1 & 0\\ 
 \hline

\end{tabular}
 \end{table}

%% file: TABLES/run-30-8-8-0-10-40-0.tex
\begin{table}
\caption{Misclustering errors for the number of clusters $k$= 8 x 8 = 64  
noise level= 0 \%, cluster radius= 10 , cluster size= 40 , 
motion freedom of clusters= 0 , minimal distance between clusters: 5.83 *R
} \label{tab:8-8-0-10-40-0} 
\begin{tabular}{|l|r|r|r|r|r|r|}
\hline
& \totWithinSS &sd& \wrongClustersPerc &sd& \RelTotWithinSS &sd\\ 
 \hline
\riKmeans& 839627.6 & 113618.2& 28.23 & 4.11& 10.12 & 1.41\\ 
 \hline
\tcKmeans& 83059.99 & 1565.39& 0 & 0& 1 & 0\\ 
 \hline
\mdKmeans& 83059.99 & 1565.39& 0 & 0& 1 & 0\\ 
 \hline
\glKmeans& 83059.99 & 1565.39& 0 & 0& 1 & 0\\ 
 \hline
\ppKmeans& 269336.8 & 139768.3& 7.19 & 5.38& 3.24 & 1.67\\ 
 \hline
\pbKmeans& 83059.99 & 1565.39& 0 & 0& 1 & 0\\ 
 \hline

\end{tabular}
 \end{table}

%% file: TABLES/run-30-9-9-0-10-40-0.tex
\begin{table}
\caption{Misclustering errors for the number of clusters $k$= 9 x 9 = 81  
noise level= 0 \%, cluster radius= 10 , cluster size= 40 , 
motion freedom of clusters= 0 , minimal distance between clusters: 5.83 *R
} \label{tab:9-9-0-10-40-0} 
\begin{tabular}{|l|r|r|r|r|r|r|}
\hline
& \totWithinSS &sd& \wrongClustersPerc &sd& \RelTotWithinSS &sd\\ 
 \hline
\riKmeans& 1121321 & 115220.7& 29.55 & 2.97& 10.67 & 1.1\\ 
 \hline
\tcKmeans& 105069.2 & 1628.07& 0 & 0& 1 & 0\\ 
 \hline
\mdKmeans& 105069.2 & 1628.07& 0 & 0& 1 & 0\\ 
 \hline
\glKmeans& 105069.2 & 1628.07& 0 & 0& 1 & 0\\ 
 \hline
\ppKmeans& 334273.7 & 110026.9& 7.04 & 3.36& 3.18 & 1.04\\ 
 \hline
\pbKmeans& 105069.2 & 1628.07& 0 & 0& 1 & 0\\ 
 \hline

\end{tabular}
 \end{table}

%% file: TABLES/run-30-10-10-0-10-40-0.tex
\begin{table}
\caption{Misclustering errors for the number of clusters $k$= 10 x 10 = 100  
noise level= 0 \%, cluster radius= 10 , cluster size= 40 , 
motion freedom of clusters= 0 , minimal distance between clusters: 5.83 *R
} \label{tab:10-10-0-10-40-0} 
\begin{tabular}{|l|r|r|r|r|r|r|}
\hline
& \totWithinSS &sd& \wrongClustersPerc &sd& \RelTotWithinSS &sd\\ 
 \hline
\riKmeans& 1452519 & 133942.8& 30.93 & 3.26& 11.2 & 0.94\\ 
 \hline
\tcKmeans& 129623.5 & 2345.05& 0 & 0& 1 & 0\\ 
 \hline
\mdKmeans& 129623.5 & 2345.05& 0 & 0& 1 & 0\\ 
 \hline
\glKmeans& 129623.5 & 2345.05& 0 & 0& 1 & 0\\ 
 \hline
\ppKmeans& 449299.6 & 148633.3& 7.9 & 3.67& 3.47 & 1.15\\ 
 \hline
\pbKmeans& 129623.5 & 2345.05& 0 & 0& 1 & 0\\ 
 \hline

\end{tabular}
 \end{table}

%% file: PDFS/worst_riKmeans_10-10-0-10-40-0.tex
\begin{figure}
\begin{center}
\includegraphics[width=0.49\textwidth]{ 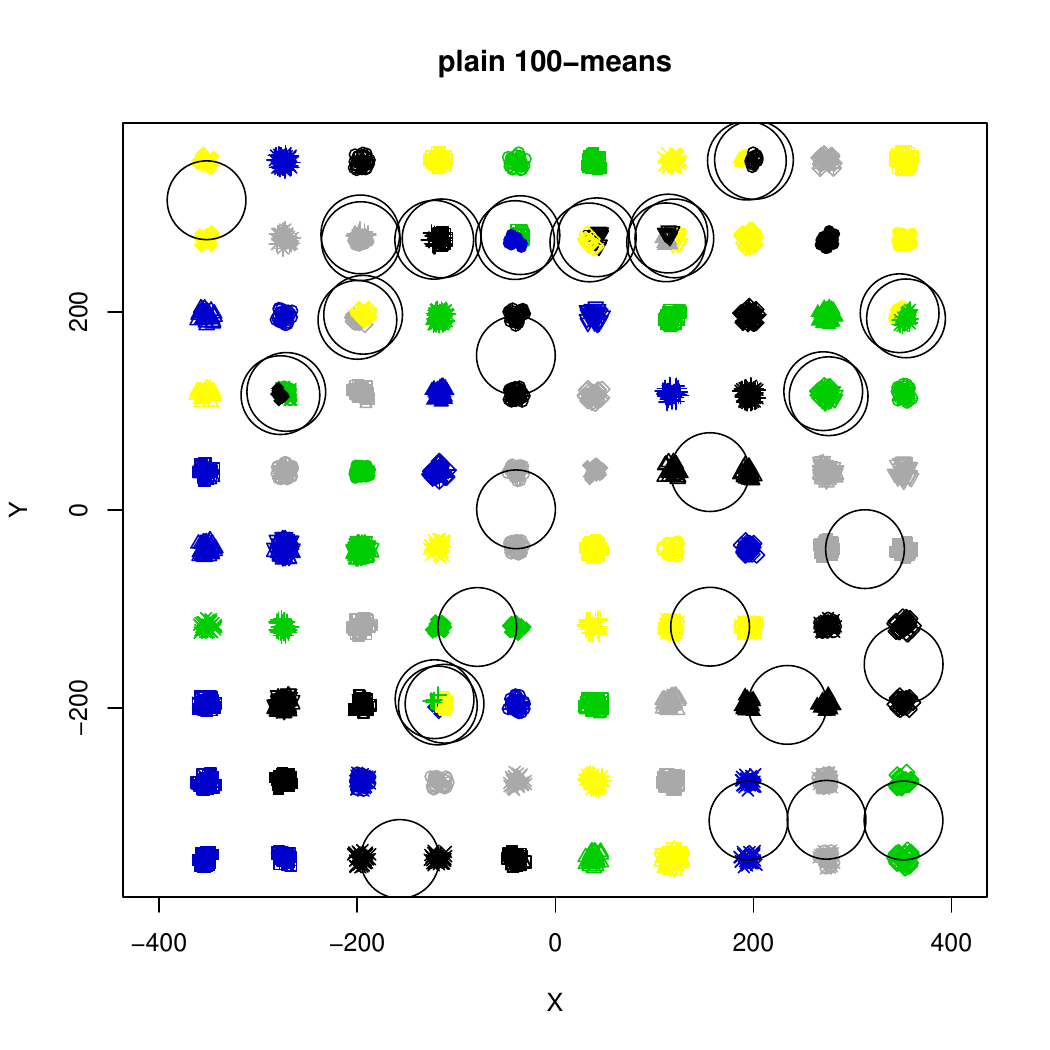 } %
\end{center}
\caption{Worst misclustering errors
 \riKmeans {} 
 for the number of clusters $k$= 10 x 10 = 100  
noise level= 0 \%, cluster radius= 10 , cluster size= 40 , 
motion freedom of clusters= 0 , 
minimal distance between clusters: 5.83 *R
} \label{fig:riKmeans-10-10-0-10-40-0} 
 \end{figure}

%% file: PDFS/worst_ppKmeans_10-10-0-10-40-0.tex
\begin{figure}
\begin{center}
\includegraphics[width=0.49\textwidth]{ 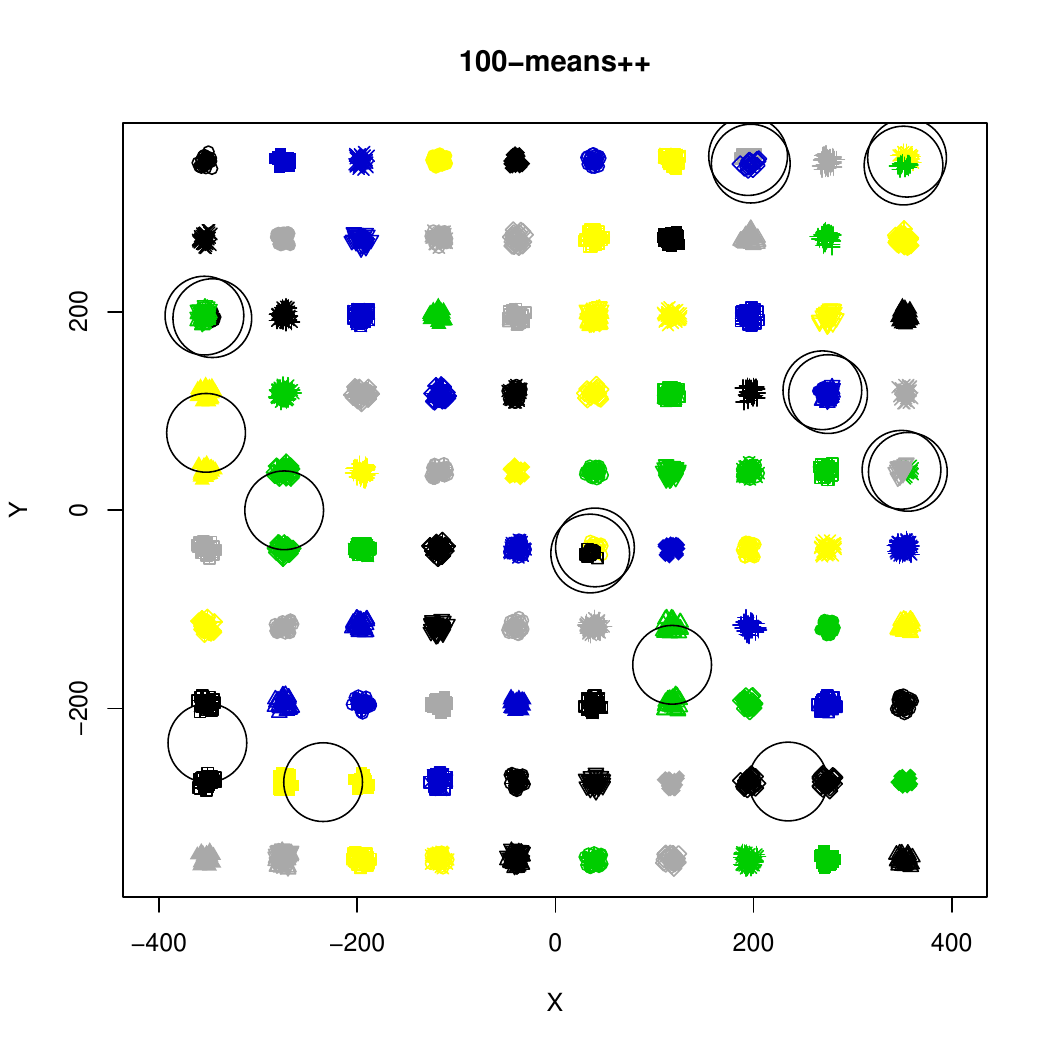 } %
\end{center}
\caption{Worst misclustering errors
 \ppKmeans {} 
 for the number of clusters $k$= 10 x 10 = 100  
noise level= 0 \%, cluster radius= 10 , cluster size= 40 , 
motion freedom of clusters= 0 , 
minimal distance between clusters: 5.83 *R
} \label{fig:ppKmeans-10-10-0-10-40-0} 
 \end{figure}

%% file: TABLES/run-30-8-8-10-10-40-0.tex
\begin{table}
\caption{Misclustering errors for the number of clusters $k$= 8 x 8 = 64  
noise level= 10 \%, cluster radius= 10 , cluster size= 40 , 
motion freedom of clusters= 0 , minimal distance between clusters: 5.83 *R} \label{tab:8-8-10-10-40-0} 
\begin{tabular}{|l|r|r|r|r|r|r|}
\hline
& \totWithinSS &sd& \wrongClustersPerc &sd& \RelTotWithinSS &sd\\ 
 \hline
\riKmeans& 898477.4 & 107564.2& 24.93 & 3.84& 10.59 & 1.29\\ 
 \hline
\tcKmeans& 84862.53 & 1680.56& 0 & 0& 1 & 0\\ 
 \hline
\mdKmeans& 92855.78 & 29994.68& 0.26 & 1& 1.09 & 0.35\\ 
 \hline
\glKmeans& 206743.7 & 3301.28& 3.17 & NaN& 2.44 & 0.04\\ 
 \hline
\ppKmeans& 352599 & 141800& 9.98 & 5.45& 4.16 & 1.7\\ 
 \hline
\pbKmeans& 84863.96 & 1681.49& 0 & 0& 1 & 0\\ 
 \hline

\end{tabular}
 \end{table}

%% file: TABLES/run-30-8-8-20-10-40-0.tex
\begin{table}
\caption{Misclustering errors for the number of clusters $k$= 8 x 8 = 64  
noise level= 20 \%, cluster radius= 10 , cluster size= 40 , 
motion freedom of clusters= 0 , minimal distance between clusters: 5.83 *R} \label{tab:8-8-20-10-40-0} 
\begin{tabular}{|l|r|r|r|r|r|r|}
\hline
& \totWithinSS &sd& \wrongClustersPerc &sd& \RelTotWithinSS &sd\\ 
 \hline
\riKmeans& 978119.1 & 132268.8& 23.56 & 3.56& 11.48 & 1.54\\ 
 \hline
\tcKmeans& 85174.19 & 1377.86& 0 & 0& 1 & 0\\ 
 \hline
\mdKmeans& 93401.87 & 30463.09& 0.31 & 1.17& 1.1 & 0.37\\ 
 \hline
\glKmeans& 204133.2 & 22354.87& 3.06 & 0.57& 2.4 & 0.26\\ 
 \hline
\ppKmeans& 348469.3 & 112942.2& 9.64 & 4.26& 4.1 & 1.34\\ 
 \hline
\pbKmeans& 85175.5 & 1376.77& 0 & 0& 1 & 0\\ 
 \hline

\end{tabular}
 \end{table}

%% file: TABLES/run-30-8-8-30-10-40-0.tex
\begin{table}
\caption{Misclustering errors for the number of clusters $k$= 8 x 8 = 64  
noise level= 30 \%, cluster radius= 10 , cluster size= 40 , 
motion freedom of clusters= 0 , minimal distance between clusters: 5.83 *R} \label{tab:8-8-30-10-40-0} 
\begin{tabular}{|l|r|r|r|r|r|r|}
\hline
& \totWithinSS &sd& \wrongClustersPerc &sd& \RelTotWithinSS &sd\\ 
 \hline
\riKmeans& 1080724 & 99859.57& 23.15 & 2.66& 12.69 & 1.27\\ 
 \hline
\tcKmeans& 85278.66 & 1548.65& 0 & 0& 1 & 0\\ 
 \hline
\mdKmeans& 134501 & 75551.75& 1.58 & 2.47& 1.57 & 0.88\\ 
 \hline
\glKmeans& 207550.5 & 3081.93& 3.17 & NaN& 2.43 & 0.04\\ 
 \hline
\ppKmeans& 416564.9 & 128271.7& 11.68 & 4.62& 4.89 & 1.5\\ 
 \hline
\pbKmeans& 85275.53 & 1542.16& 0 & 0& 1 & 0\\ 
 \hline

\end{tabular}
 \end{table}

%% file: TABLES/run-30-8-8-40-10-40-0.tex
\begin{table}
\caption{Misclustering errors for the number of clusters $k$= 8 x 8 = 64  
noise level= 40 \%, cluster radius= 10 , cluster size= 40 , 
motion freedom of clusters= 0 , minimal distance between clusters: 5.83 *R} \label{tab:8-8-40-10-40-0} 
\begin{tabular}{|l|r|r|r|r|r|r|}
\hline
& \totWithinSS &sd& \wrongClustersPerc &sd& \RelTotWithinSS &sd\\ 
 \hline
\riKmeans& 1162857 & 118303.5& 22.2 & 3.35& 13.55 & 1.37\\ 
 \hline
\tcKmeans& 85807.74 & 1429.99& 0 & 0& 1 & 0\\ 
 \hline
\mdKmeans& 135133.9 & 74748.25& 1.68 & 2.63& 1.58 & 0.87\\ 
 \hline
\glKmeans& 208272.4 & 3781.91& 3.17 & NaN& 2.43 & 0.04\\ 
 \hline
\ppKmeans& 437003.7 & 126730& 12.39 & 4.48& 5.09 & 1.46\\ 
 \hline
\pbKmeans& 85809.42 & 1430.65& 0 & 0& 1 & 0\\ 
 \hline

\end{tabular}
 \end{table}

%% file: TABLES/run-30-8-8-50-10-40-0.tex
\begin{table}
\caption{Misclustering errors for the number of clusters $k$= 8 x 8 = 64  
noise level= 50 \%, cluster radius= 10 , cluster size= 40 , 
motion freedom of clusters= 0 , minimal distance between clusters: 5.83 *R
} \label{tab:8-8-50-10-40-0} 
\begin{tabular}{|l|r|r|r|r|r|r|}
\hline
& \totWithinSS &sd& \wrongClustersPerc &sd& \RelTotWithinSS &sd\\ 
 \hline
\riKmeans& 1304727 & 119562.3& 23.05 & 3.07& 15.06 & 1.32\\ 
 \hline
\tcKmeans& 86632.72 & 1362.46& 0 & 0& 1 & 0\\ 
 \hline
\mdKmeans& 147695 & 68845.15& 2.04 & 2.43& 1.7 & 0.79\\ 
 \hline
\glKmeans& 205161.5 & 22190.22& 3.12 & 0.64& 2.37 & 0.26\\ 
 \hline
\ppKmeans& 424568.7 & 123582.4& 12.11 & 4.74& 4.91 & 1.45\\ 
 \hline
\pbKmeans& 94858.91 & 30873.22& 0.26 & 1& 1.09 & 0.35\\ 
 \hline

\end{tabular}
 \end{table}

%% file: PDFS/worst_riKmeans_8-8-30-10-40-0.tex
\begin{figure}
\begin{center}
\includegraphics[width=0.49\textwidth]{ 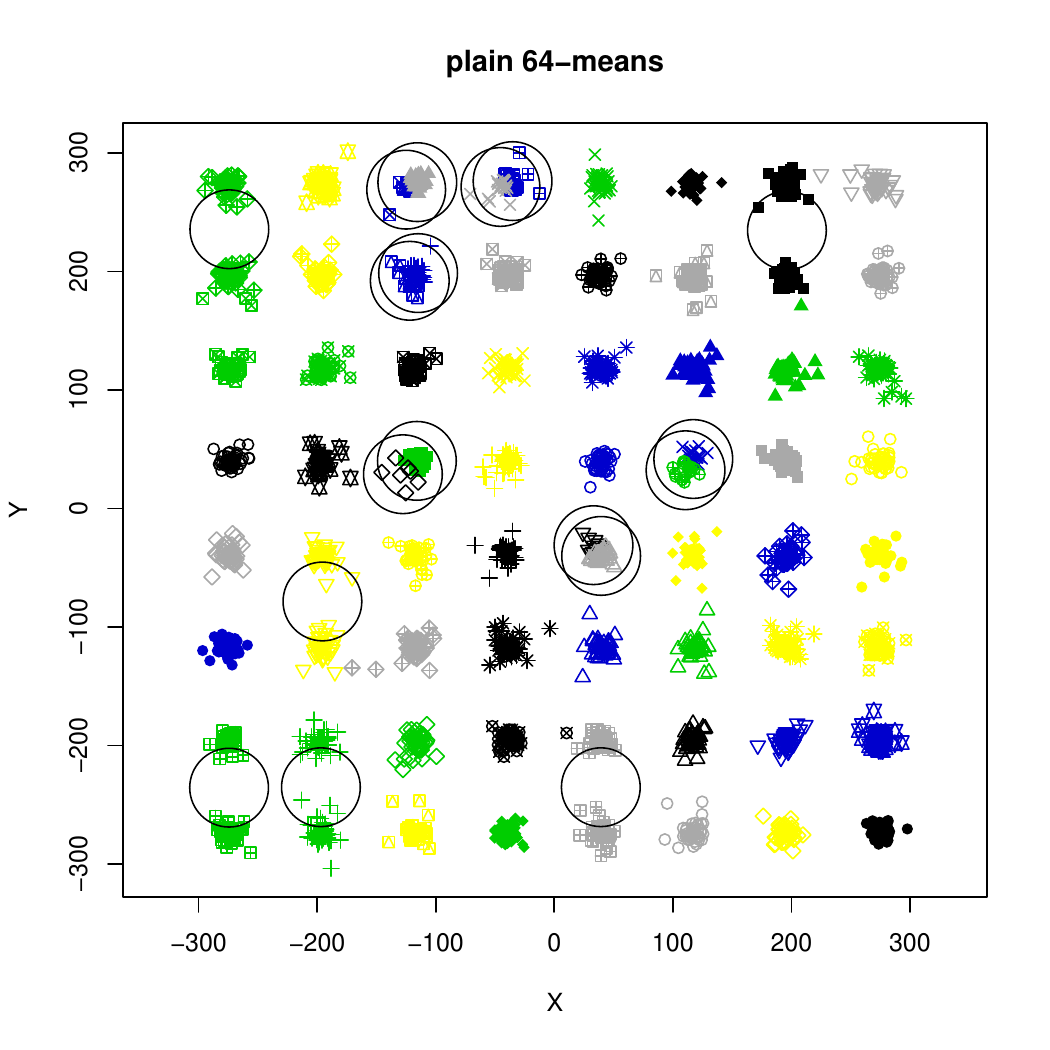 } %
\end{center}
\caption{Worst misclustering errors
 \riKmeans {} 
 for the number of clusters $k$= 8 x 8 = 64  
noise level= 30 \%, cluster radius= 10 , cluster size= 40 , 
motion freedom of clusters= 0 , 
minimal distance between clusters: 5.83 *R} \label{fig:riKmeans-8-8-30-10-40-0} 
 \end{figure}

%% file: PDFS/worst_mdKmeans_8-8-30-10-40-0.tex
\begin{figure}
\begin{center}
\includegraphics[width=0.49\textwidth]{ 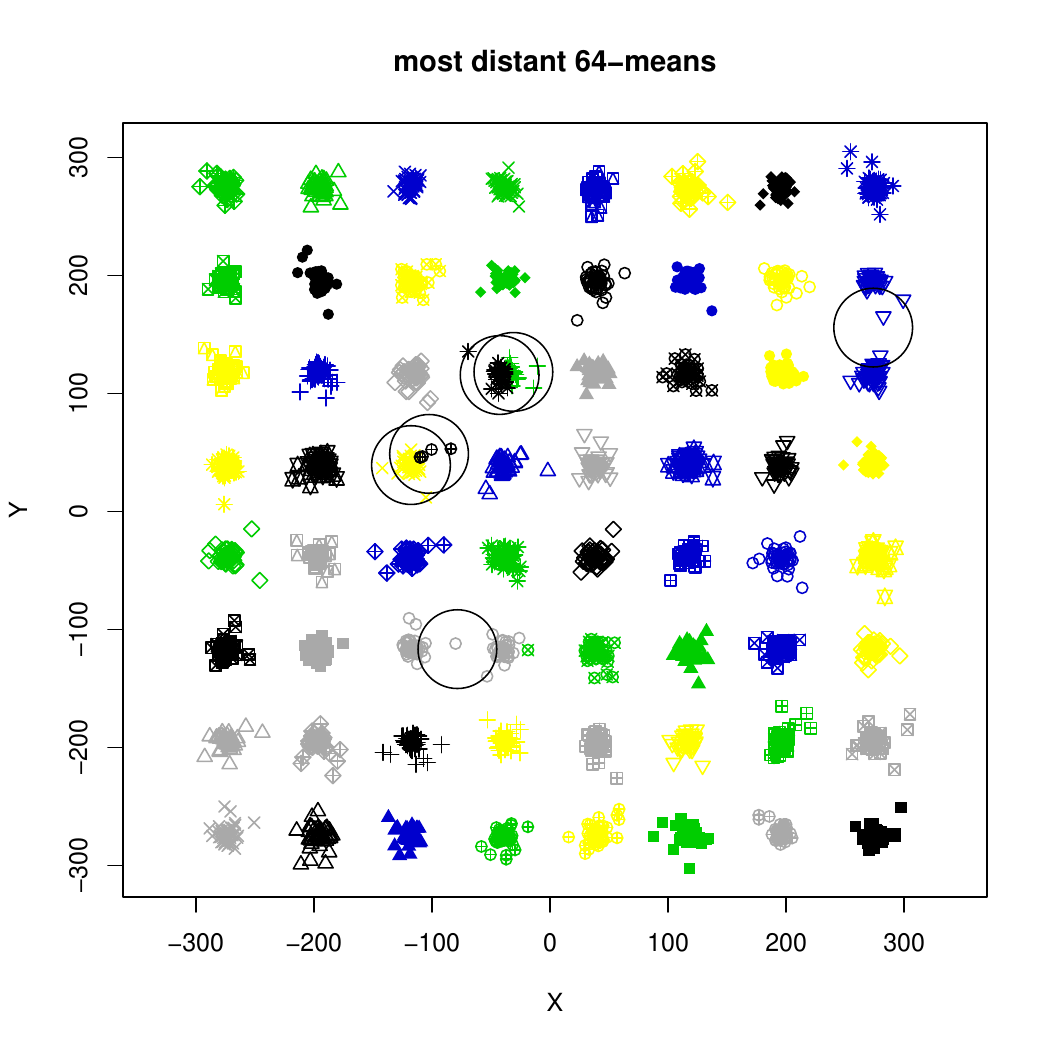 } %
\end{center}
\caption{Worst misclustering errors
 \mdKmeans {} 
 for the number of clusters $k$= 8 x 8 = 64  
noise level= 30 \%, cluster radius= 10 , cluster size= 40 , 
motion freedom of clusters= 0 , 
minimal distance between clusters: 5.83 *R} \label{fig:mdKmeans-8-8-30-10-40-0} 
 \end{figure}

%% file: PDFS/worst_ppKmeans_8-8-30-10-40-0.tex
\begin{figure}
\begin{center}
\includegraphics[width=0.49\textwidth]{ 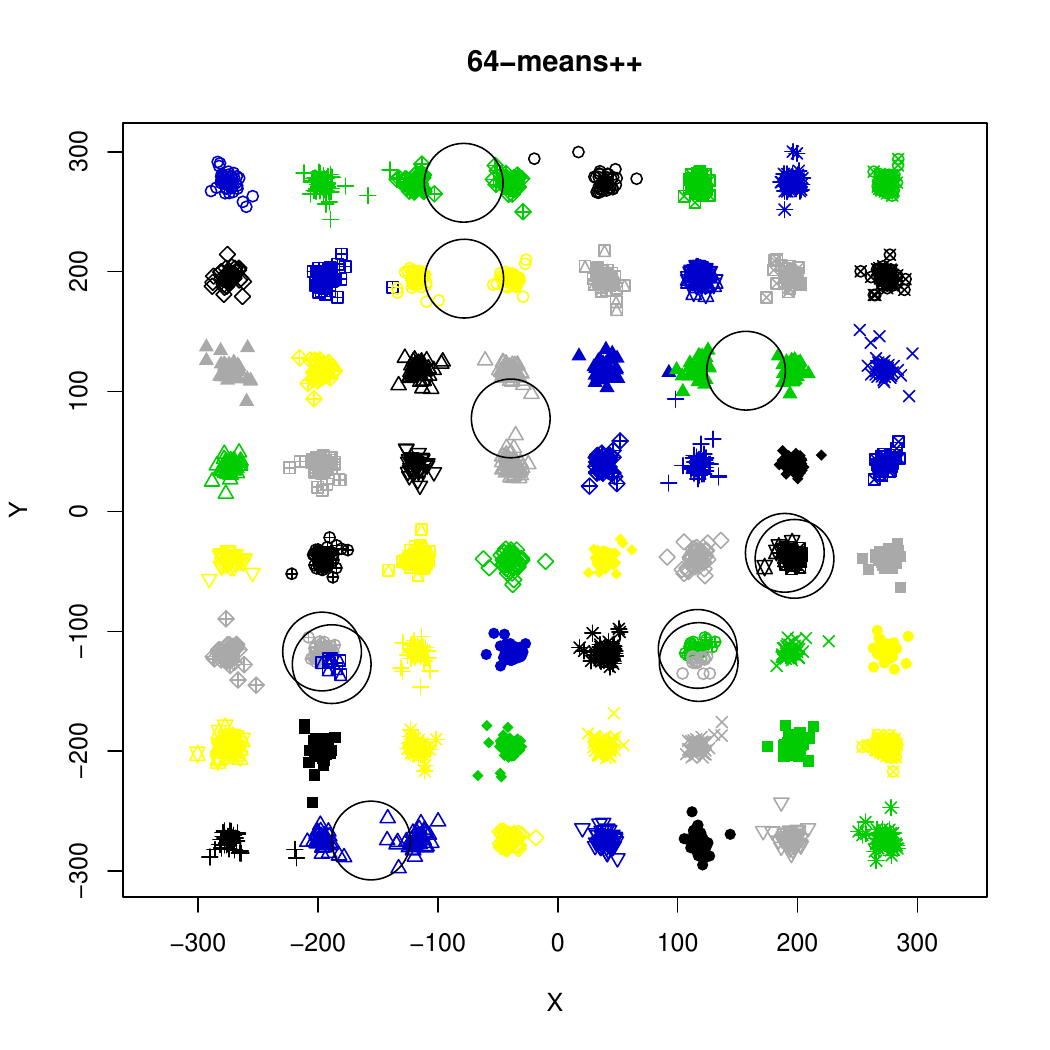 } %
\end{center}
\caption{Worst misclustering errors
 \ppKmeans {} 
 for the number of clusters $k$= 8 x 8 = 64  
noise level= 30 \%, cluster radius= 10 , cluster size= 40 , 
motion freedom of clusters= 0 , 
minimal distance between clusters: 5.83 *R} \label{fig:ppKmeans-8-8-30-10-40-0} 
 \end{figure}

%% file: PDFS/worst_glKmeans_8-8-30-10-40-0.tex
\begin{figure}
\begin{center}
\includegraphics[width=0.49\textwidth]{ 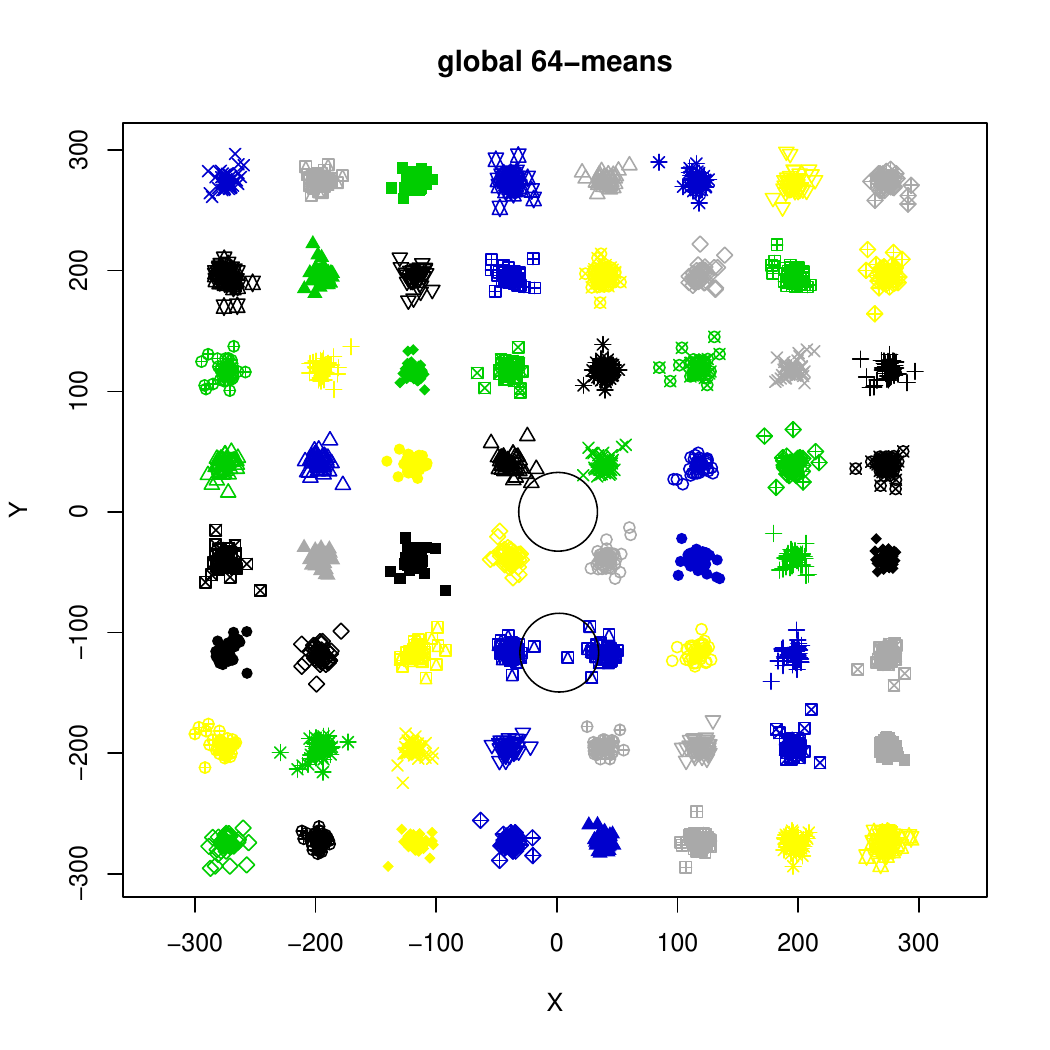 } %
\end{center}
\caption{Worst misclustering errors
 \glKmeans {} 
 for the number of clusters $k$= 8 x 8 = 64  
noise level= 30 \%, cluster radius= 10 , cluster size= 40 , 
motion freedom of clusters= 0 , 
minimal distance between clusters: 5.83 *R} \label{fig:glKmeans-8-8-30-10-40-0} 
 \end{figure}

%% file: TABLES/run-30-8-8-30-10-40-5.tex
\begin{table}
\caption{Misclustering errors for the number of clusters $k$= 8 x 8 = 64  
noise level= 30 \%, cluster radius= 10 , cluster size= 40 , 
motion freedom of clusters= 5 , minimal distance between clusters: 5.83 *R} \label{tab:8-8-30-10-40-5} 
\begin{tabular}{|l|r|r|r|r|r|r|}
\hline
& \totWithinSS &sd& \wrongClustersPerc &sd& \RelTotWithinSS &sd\\ 
 \hline
\riKmeans& 1324228 & 164858.9& 24.78 & 3.98& 15.39 & 1.93\\ 
 \hline
\tcKmeans& 86053.02 & 1553.69& 0 & 0& 1 & 0\\ 
 \hline
\mdKmeans& 95193.42 & 33978.65& 0.31 & 1.17& 1.11 & 0.4\\ 
 \hline
\glKmeans& 207401.8 & 48454.39& 2.75 & 1.08& 2.41 & 0.56\\ 
 \hline
\ppKmeans& 494330.4 & 151398.7& 12.58 & 4.8& 5.75 & 1.78\\ 
 \hline
\pbKmeans& 86053.02 & 1553.69& 0 & 0& 1 & 0\\ 
 \hline

\end{tabular}
 \end{table}

%% file: TABLES/run-30-8-8-30-10-40-10.tex
\begin{table}
\caption{Misclustering errors for the number of clusters $k$= 8 x 8 = 64  
noise level= 30 \%, cluster radius= 10 , cluster size= 40 , 
motion freedom of clusters= 10 , minimal distance between clusters: 5.83 *R} \label{tab:8-8-30-10-40-10} 
\begin{tabular}{|l|r|r|r|r|r|r|}
\hline
& \totWithinSS &sd& \wrongClustersPerc &sd& \RelTotWithinSS &sd\\ 
 \hline
\riKmeans& 1695803 & 225108.9& 28.52 & 4.04& 19.71 & 2.55\\ 
 \hline
\tcKmeans& 86016.3 & 1185.21& 0 & 0& 1 & 0\\ 
 \hline
\mdKmeans& 86016.3 & 1185.21& 0 & 0& 1 & 0\\ 
 \hline
\glKmeans& 218801.5 & 55080.32& 2.75 & 1.08& 2.54 & 0.64\\ 
 \hline
\ppKmeans& 451174.5 & 218741.7& 9.6 & 5.47& 5.25 & 2.56\\ 
 \hline
\pbKmeans& 86016.3 & 1185.21& 0 & 0& 1 & 0\\ 
 \hline

\end{tabular}
 \end{table}

%% file: TABLES/run-30-8-8-30-10-40-20.tex
\begin{table}
\caption{Misclustering errors for the number of clusters $k$= 8 x 8 = 64  
noise level= 30 \%, cluster radius= 10 , cluster size= 40 , 
motion freedom of clusters= 20 , minimal distance between clusters: 5.83 *R} \label{tab:8-8-30-10-40-20} 
\begin{tabular}{|l|r|r|r|r|r|r|}
\hline
& \totWithinSS &sd& \wrongClustersPerc &sd& \RelTotWithinSS &sd\\ 
 \hline
\riKmeans& 2510247 & 271035& 35.27 & 3.71& 29.36 & 3.18\\ 
 \hline
\tcKmeans& 85506.64 & 1525.18& 0 & 0& 1 & 0\\ 
 \hline
\mdKmeans& 85506.64 & 1525.18& 0 & 0& 1 & 0\\ 
 \hline
\glKmeans& 197579.5 & 75928.16& 2.22 & 1.45& 2.31 & 0.89\\ 
 \hline
\ppKmeans& 491747 & 241062.7& 9.14 & 5.64& 5.75 & 2.8\\ 
 \hline
\pbKmeans& 85506.64 & 1525.18& 0 & 0& 1 & 0\\ 
 \hline

\end{tabular}
 \end{table}

%% file: TABLES/run-30-8-8-30-10-40-40.tex
\begin{table}
\caption{Misclustering errors for the number of clusters $k$= 8 x 8 = 64  
noise level= 30 \%, cluster radius= 10 , cluster size= 40 , 
motion freedom of clusters= 40 , minimal distance between clusters: 5.83 *R} \label{tab:8-8-30-10-40-40} 
\begin{tabular}{|l|r|r|r|r|r|r|}
\hline
& \totWithinSS &sd& \wrongClustersPerc &sd& \RelTotWithinSS &sd\\ 
 \hline
\riKmeans& 4116153 & 479531.9& 38.98 & 5.19& 48.05 & 5.75\\ 
 \hline
\tcKmeans& 85707.34 & 1508.97& 0 & 0& 1 & 0\\ 
 \hline
\mdKmeans& 85707.34 & 1508.97& 0 & 0& 1 & 0\\ 
 \hline
\glKmeans& 230218.6 & 85552.31& 2.43 & 1.34& 2.68 & 0.99\\ 
 \hline
\ppKmeans& 470169.7 & 239104.3& 6.89 & 3.79& 5.48 & 2.75\\ 
 \hline
\pbKmeans& 85707.34 & 1508.97& 0 & 0& 1 & 0\\ 
 \hline

\end{tabular}
 \end{table}

%% file: PDFS/worst_riKmeans_8-8-30-10-40-40.tex
\begin{figure}
\begin{center}
\includegraphics[width=0.49\textwidth]{ 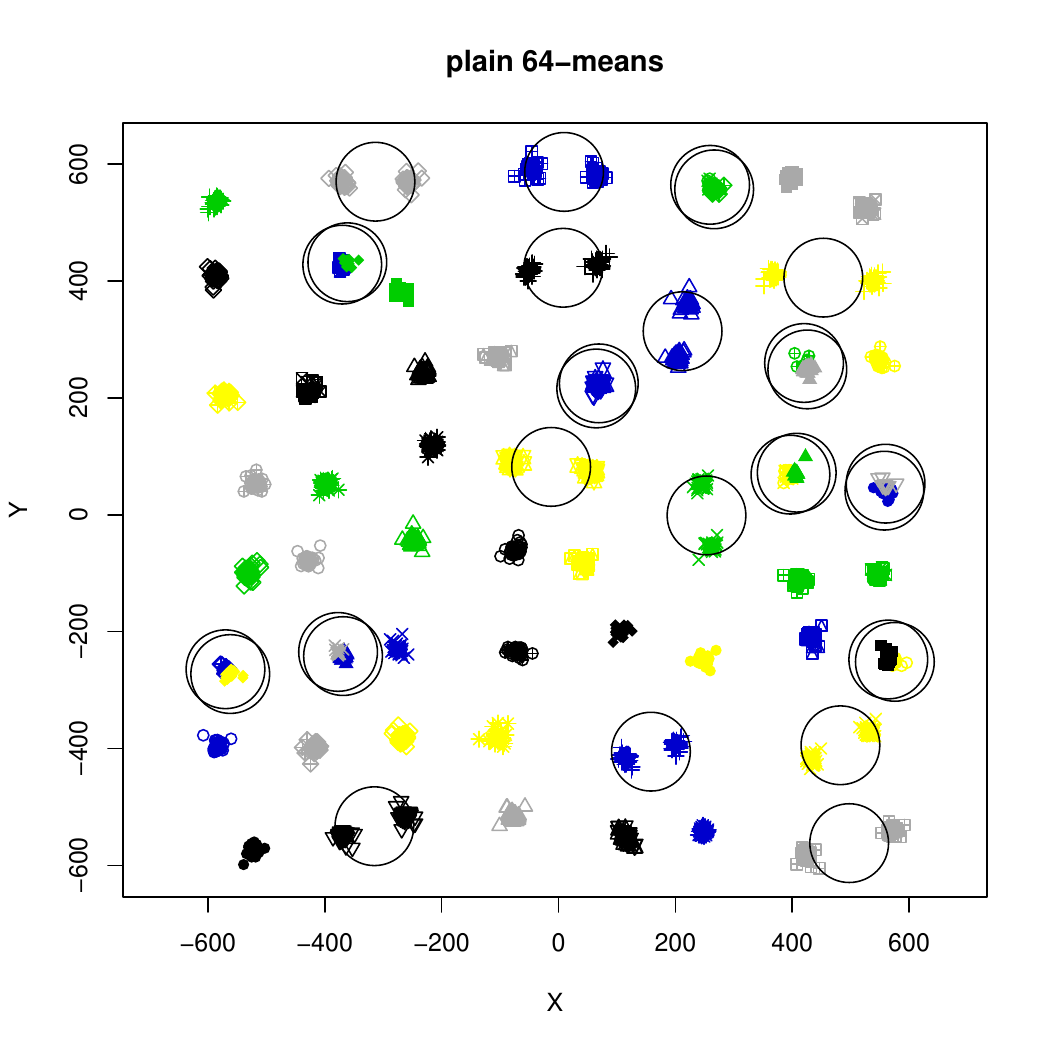 } %
\end{center}
\caption{Worst misclustering errors
 \riKmeans {} 
 for the number of clusters $k$= 8 x 8 = 64  
noise level= 30 \%, cluster radius= 10 , cluster size= 40 , 
motion freedom of clusters= 40 , 
minimal distance between clusters: 5.83 *R} \label{fig:riKmeans-8-8-30-10-40-40} 
 \end{figure}

%% file: PDFS/worst_ppKmeans_8-8-30-10-40-40.tex
\begin{figure}
\begin{center}
\includegraphics[width=0.49\textwidth]{ 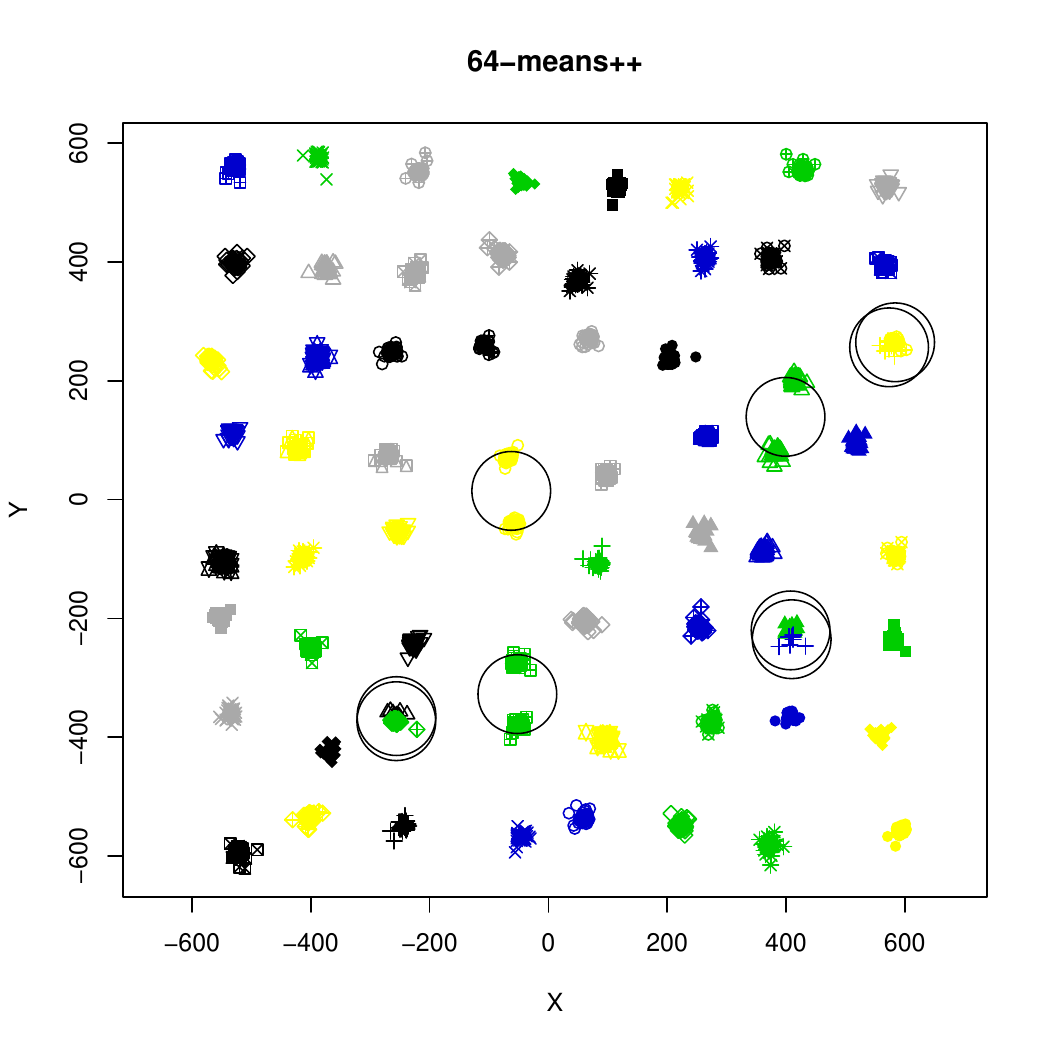 } %
\end{center}
\caption{Worst misclustering errors
 \ppKmeans {} 
 for the number of clusters $k$= 8 x 8 = 64  
noise level= 30 \%, cluster radius= 10 , cluster size= 40 , 
motion freedom of clusters= 40 , 
minimal distance between clusters: 5.83 *R} \label{fig:ppKmeans-8-8-30-10-40-40} 
 \end{figure}

%% file: PDFS/worst_glKmeans_8-8-30-10-40-40.tex
\begin{figure}
\begin{center}
\includegraphics[width=0.49\textwidth]{ 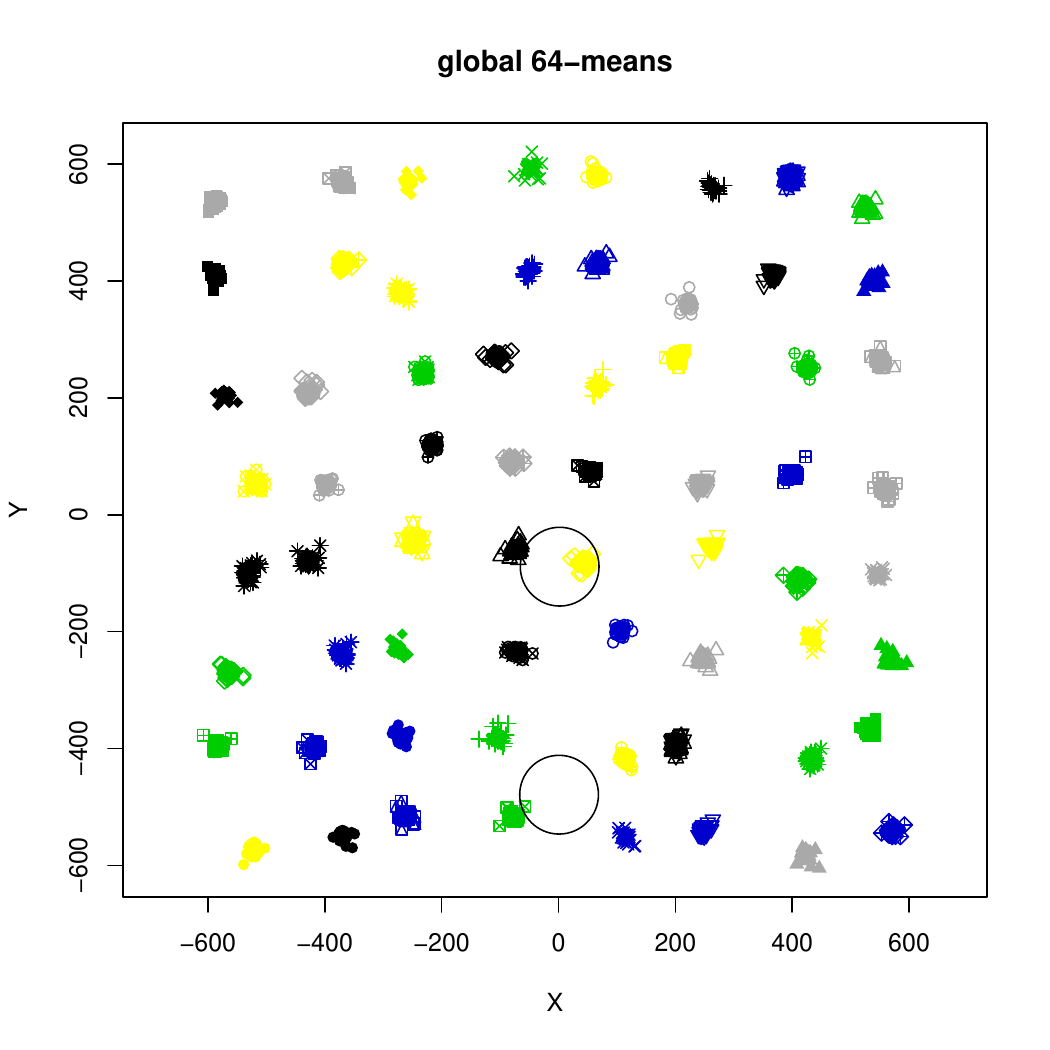 } %
\end{center}
\caption{Worst misclustering errors
 \glKmeans {} 
 for the number of clusters $k$= 8 x 8 = 64  
noise level= 30 \%, cluster radius= 10 , cluster size= 40 , 
motion freedom of clusters= 40 , 
minimal distance between clusters: 5.83 *R} \label{fig:glKmeans-8-8-30-10-40-40} 
 \end{figure}

%% file: TABLES/run-30-8-8-30-10-12-10.tex
\begin{table}
\caption{Misclustering errors for the number of clusters $k$= 8 x 8 = 64  
noise level= 30 \%, cluster radius= 10 , cluster size= 12 , 
motion freedom of clusters= 10 , minimal distance between clusters: 5.83 *R} \label{tab:8-8-30-10-12-10} 
\begin{tabular}{|l|r|r|r|r|r|r|}
\hline
& \totWithinSS &sd& \wrongClustersPerc &sd& \RelTotWithinSS &sd\\ 
 \hline
\riKmeans& 532774.8 & 59062.4& 26.9 & 4.59& 20.21 & 2.32\\ 
 \hline
\tcKmeans& 26384.5 & 888.86& 0 & 0& 1 & 0\\ 
 \hline
\mdKmeans& 26384.5 & 888.86& 0 & 0& 1 & 0\\ 
 \hline
\glKmeans& 56905.01 & 20166.2& 2.22 & 1.45& 2.17 & 0.78\\ 
 \hline
\ppKmeans& 152971.8 & 48444.23& 10.39 & 4.27& 5.8 & 1.83\\ 
 \hline
\pbKmeans& 26384.5 & 888.86& 0 & 0& 1 & 0\\ 
 \hline

\end{tabular}
 \end{table}

%% file: TABLES/run-30-8-8-30-10-24-10.tex
\begin{table}
\caption{Misclustering errors for the number of clusters $k$= 8 x 8 = 64  
noise level= 30 \%, cluster radius= 10 , cluster size= 24 , 
motion freedom of clusters= 10 , minimal distance between clusters: 5.83 *R} \label{tab:8-8-30-10-24-10} 
\begin{tabular}{|l|r|r|r|r|r|r|}
\hline
& \totWithinSS &sd& \wrongClustersPerc &sd& \RelTotWithinSS &sd\\ 
 \hline
\riKmeans& 1036702 & 92429.47& 29.33 & 3.53& 20.1 & 1.78\\ 
 \hline
\tcKmeans& 51597.87 & 965.31& 0 & 0& 1 & 0\\ 
 \hline
\mdKmeans& 51596.23 & 961.13& 0 & 0& 1 & 0\\ 
 \hline
\glKmeans& 117913.5 & 40946.04& 2.32 & 1.4& 2.28 & 0.79\\ 
 \hline
\ppKmeans& 255396.2 & 101449.6& 8.32 & 4.28& 4.95 & 1.95\\ 
 \hline
\pbKmeans& 51596.23 & 961.13& 0 & 0& 1 & 0\\ 
 \hline

\end{tabular}
 \end{table}

%% file: TABLES/run-30-8-8-30-10-48-10.tex
\begin{table}
\caption{Misclustering errors for the number of clusters $k$= 8 x 8 = 64  
noise level= 30 \%, cluster radius= 10 , cluster size= 48 , 
motion freedom of clusters= 10 , minimal distance between clusters: 5.83 *R} \label{tab:8-8-30-10-48-10} 
\begin{tabular}{|l|r|r|r|r|r|r|}
\hline
& \totWithinSS &sd& \wrongClustersPerc &sd& \RelTotWithinSS &sd\\ 
 \hline
\riKmeans& 2032025 & 191414.4& 28.58 & 3.33& 19.77 & 1.92\\ 
 \hline
\tcKmeans& 102804.7 & 1817.95& 0 & 0& 1 & 0\\ 
 \hline
\mdKmeans& 113691.9 & 40514.85& 0.26 & 1& 1.11 & 0.4\\ 
 \hline
\glKmeans& 254761.8 & 60728.79& 2.75 & 1.08& 2.48 & 0.6\\ 
 \hline
\ppKmeans& 514296.1 & 222361.9& 9.49 & 5.27& 5 & 2.14\\ 
 \hline
\pbKmeans& 102804.7 & 1817.95& 0 & 0& 1 & 0\\ 
 \hline

\end{tabular}
 \end{table}

%% file: TABLES/run-30-8-8-30-10-96-10.tex
\begin{table}
\caption{Misclustering errors for the number of clusters $k$= 8 x 8 = 64  
noise level= 30 \%, cluster radius= 10 , cluster size= 96 , 
motion freedom of clusters= 10 , minimal distance between clusters: 5.83 *R} \label{tab:8-8-30-10-96-10} 
\begin{tabular}{|l|r|r|r|r|r|r|}
\hline
& \totWithinSS &sd& \wrongClustersPerc &sd& \RelTotWithinSS &sd\\ 
 \hline
\riKmeans& 4169196 & 357641.3& 29.67 & 3.49& 20.34 & 1.68\\ 
 \hline
\tcKmeans& 204907 & 2139.58& 0 & 0& 1 & 0\\ 
 \hline
\mdKmeans& 228679.3 & 89152.12& 0.31 & 1.17& 1.12 & 0.43\\ 
 \hline
\glKmeans& 555050 & 72818.57& 3.06 & 0.57& 2.71 & 0.36\\ 
 \hline
\ppKmeans& 1082762 & 418762.2& 10.37 & 4.83& 5.28 & 2.03\\ 
 \hline
\pbKmeans& 204907 & 2139.58& 0 & 0& 1 & 0\\ 
 \hline

\end{tabular}
 \end{table}

%% file: PDFS/worst_riKmeans_8-8-30-10-96-10.tex
\begin{figure}
\begin{center}
\includegraphics[width=0.49\textwidth]{ 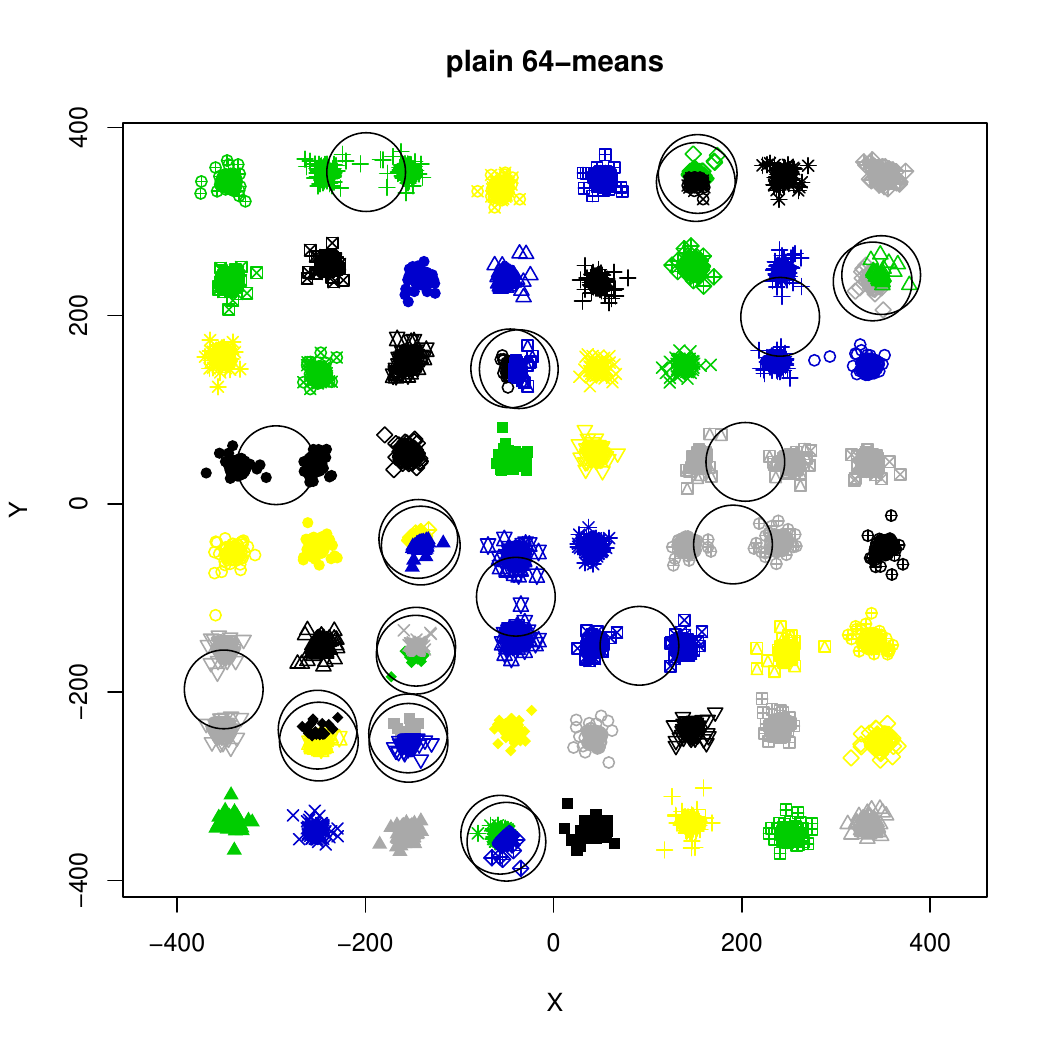 } %
\end{center}
\caption{Worst misclustering errors
 \riKmeans {} 
 for the number of clusters $k$= 8 x 8 = 64  
noise level= 30 \%, cluster radius= 10 , cluster size= 96 , 
motion freedom of clusters= 10 , 
minimal distance between clusters: 5.83 *R} \label{fig:riKmeans-8-8-30-10-96-10} 
 \end{figure}

%% file: PDFS/worst_ppKmeans_8-8-30-10-96-10.tex
\begin{figure}
\begin{center}
\includegraphics[width=0.49\textwidth]{ 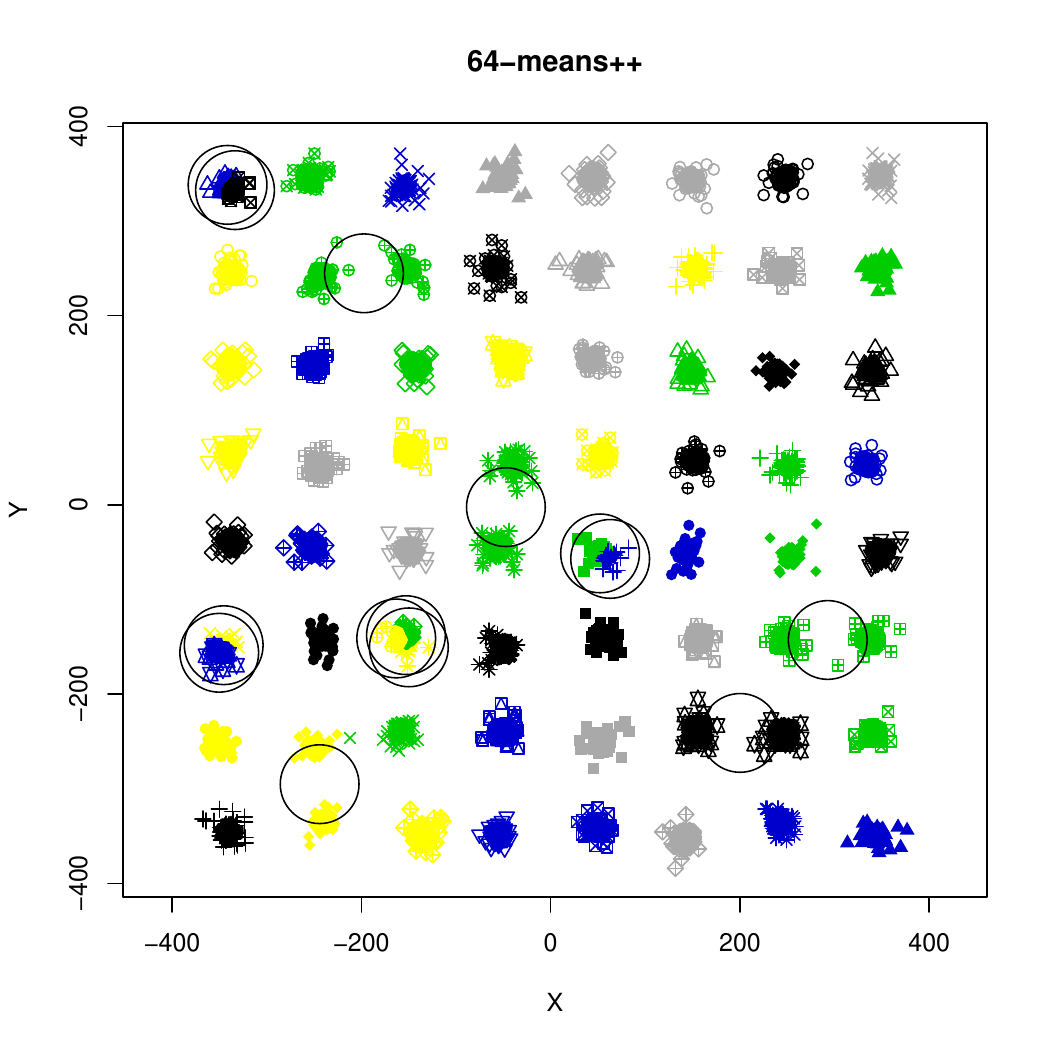 } %
\end{center}
\caption{Worst misclustering errors
 \ppKmeans {} 
 for the number of clusters $k$= 8 x 8 = 64  
noise level= 30 \%, cluster radius= 10 , cluster size= 96 , 
motion freedom of clusters= 10 , 
minimal distance between clusters: 5.83 *R} \label{fig:ppKmeans-8-8-30-10-96-10} 
 \end{figure}

%% file: PDFS/worst_glKmeans_8-8-30-10-96-10.tex
\begin{figure}
\begin{center}
\includegraphics[width=0.49\textwidth]{ 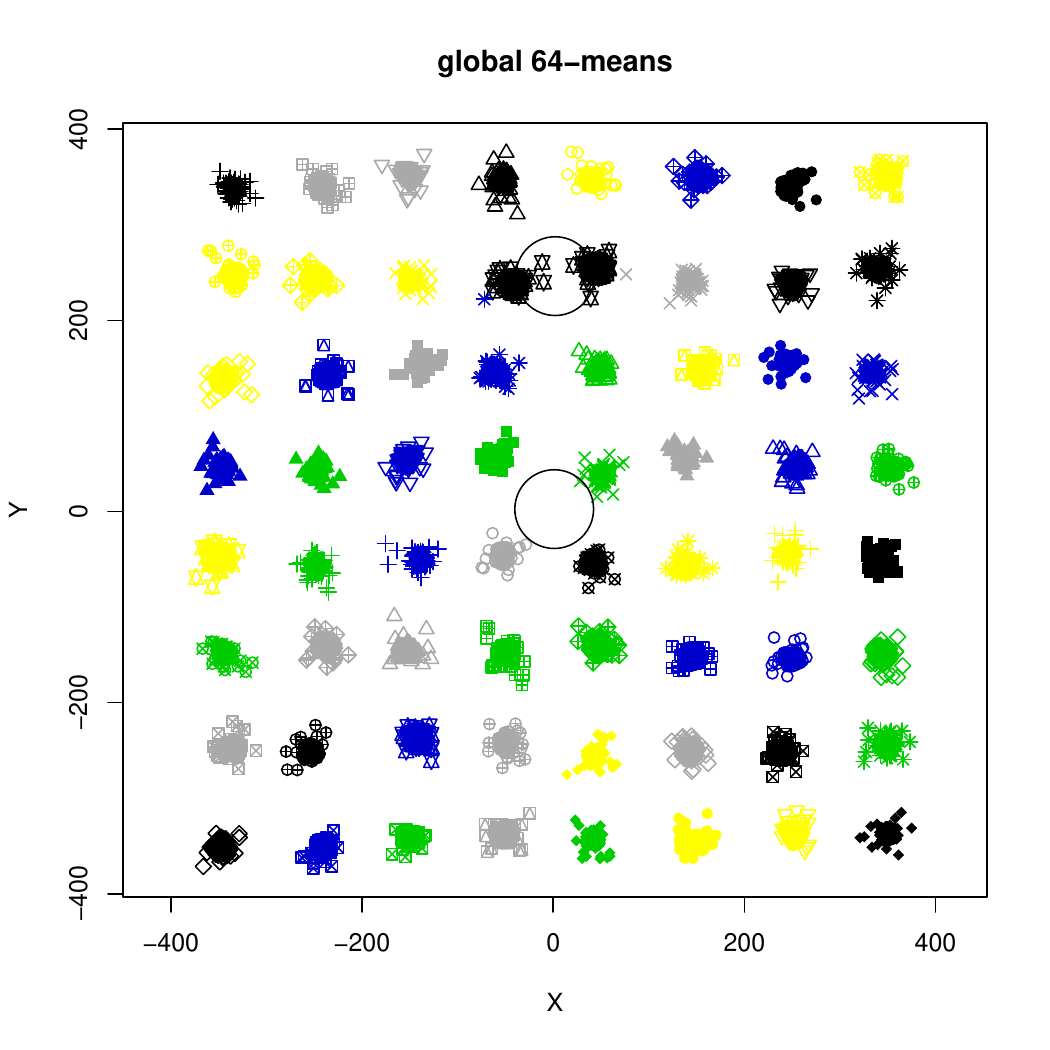 } %
\end{center}
\caption{Worst misclustering errors
 \glKmeans {} 
 for the number of clusters $k$= 8 x 8 = 64  
noise level= 30 \%, cluster radius= 10 , cluster size= 96 , 
motion freedom of clusters= 10 , 
minimal distance between clusters: 5.83 *R} \label{fig:glKmeans-8-8-30-10-96-10} 
 \end{figure}